\pdfoutput=1

\documentclass[11pt]{article}

\usepackage{acl}

\usepackage{times}
\usepackage{latexsym}
\usepackage{graphicx}
\usepackage{amsmath,bm}
\usepackage{amssymb}
\usepackage{multirow}
\usepackage{xcolor}

\usepackage[T1]{fontenc}

\usepackage[utf8]{inputenc}

\usepackage{microtype}

\title{Efficient Joint Learning for Clinical Named Entity Recognition and Relation Extraction Using Fourier Networks: A Use Case in Adverse Drug Events}

\author{Anthony Yazdani, Dimitrios Proios, Hossein Rouhizadeh, Douglas Teodoro \\
  University of Geneva, Faculty of medicine,\\
  Department of radiology and medical informatics, Data science for digital health\\
  firstname.lastname@unige.ch \\}

\begin{document}
\maketitle
\begin{abstract}
Current approaches for clinical information extraction are inefficient in terms of computational costs and memory consumption, hindering their application to process large-scale electronic health records (EHRs).
We propose an efficient end-to-end model, the Joint-NER-RE-Fourier (JNRF), to jointly learn the tasks of named entity recognition and relation extraction for documents of variable length. The architecture uses positional encoding and unitary batch sizes to process variable length documents and uses a weight-shared Fourier network layer for low-complexity token mixing. Finally, we reach the theoretical computational complexity lower bound for relation extraction using a selective pooling strategy and distance-aware attention weights with trainable polynomial distance functions.
We evaluated the JNRF architecture using the 2018 N2C2 ADE benchmark to jointly extract medication-related entities and relations in variable-length EHR summaries. JNRF outperforms rolling window BERT with selective pooling by 0.42\%, while being twice as fast to train. Compared to state-of-the-art BiLSTM-CRF architectures on the N2C2 ADE benchmark, results show that the proposed approach trains 22 times faster and reduces GPU memory consumption by 1.75 folds, with a reasonable performance tradeoff of 90\%, without the use of external tools, hand-crafted rules or post-processing.
Given the significant carbon footprint of deep learning models and the current energy crises, these methods could support efficient and cleaner information extraction in EHRs and other types of large-scale document databases.
\end{abstract}

\section{Introduction}
Adverse drug events (ADEs) are defined as any injury resulting from medication use and comprise the largest category of adverse events \cite{leape1991nature, bates1995incidence}. Serious ADEs have been estimated to cost from \$30 to \$137 billion in ambulatory settings in the US \cite{johnson1996drug}, and their costs have been doubling since then \cite{ernst2001drug}. Due to safety concerns, between 21\% to 27\% of marketed drugs in the US have received black-box warnings or have been withdrawn by the Food and Drug Administration (FDA) within the first 16 years of marketing \cite{frank2014era}.

Clinical notes stored in electronic health record (EHRs) systems are a valuable source of information for pharmacovigilance \cite{boland2015all}. However, only 1\% of ADEs recorded in EHRs are reported to ADE registries, such as the FDA Adverse Event Reporting System (FAERS), while coded diagnoses have low sensitivity for ADEs \cite{nadkarni2010drug, classen2011global}. Recognizing medication-related entities in clinical notes, extracting relations among them, and structuring this information can help identify ADEs in early stages of the drug marketing process, thus improving patient safety \cite{luo2017natural}. 

The state-of-the-art for biomedical named entity recognition (NER) and relation extraction (RE) is dominated by bidirectional LSTM \cite{hochreiter1997long} or BERT \cite{devlin2018bert} architectures, combined with a CRF \cite{lafferty2001conditional} layer and often hand-crafted rules \cite{xu2017uth_ccb, christopoulou2020adverse, wei2020study, henry20202018, fang2021joint}. Despite the high performance of end-to-end (E2E) NER+RE models, they have some important limitations imposed by the model complexity, e.g., quadratic in terms of entity types in the CRF layer or in terms of tokens in the dot-product attention mechanisms \cite{sutton2012introduction, shen2021efficient}, which hinders their effective application in the biomedical domain due to its large number of entities and large size of free text databases. 

A particularity of NER and RE  for pharmacovigilance is that efficient recall of entities and relations is of utmost importance, as we would like to avoid missing a serious ADE. Nevertheless, current approaches tend to automatically discard long distance (or inter-passage) relations \cite{yao2019docred, christopoulou2020adverse}. Moreover, EHR documents varies significantly in length, containing from a few hundred tokens for simpler patient records up to several thousand tokens for more complex patients (e.g., chronic diseases) \cite{henry20202018}. Due to their computational complexity, these methods cannot process EHRs in their integrity without resorting to impractical and/or inefficient techniques such as windowing strategies \cite{NEURIPS2020_96671501, pappagari2019hierarchical, yangetal2016hierarchical}. 

Ongoing research is predominantly performance-driven, leading to a resurgence of resource-intensive models, neglecting the carbon footprint of deep learning models in favor of often marginal improvement in effectiveness \cite{wei2020study, knafou2020bitem, copara2020contextualized, copara2020named, fang2021joint, naderi2021ensemble}. As a consequence of the technical constraints induced by highly complex models, these methods are currently being associated to a significant excess on carbon emissions \cite{gibney2022shrink}. The most direct impact of training and deploying a machine learning model is the emission of greenhouse gases due to the increased hardware energy consumption \cite{ligozat2021practical}. Therefore, a direct way to reduce the ecological impact of training and deploying machine learning models is to reduce the training and inference time, i.e., providing the community with low memory and computational cost models.

To tackle these limitations and issues, we propose the Joint-NER-RE-Fourier (JNRF) model with a reduced algorithmic complexity for information extraction. We combine positional encoding with unitary batch size training so that the model processes automatically variable size EHRs with consistent performance. We use a Fourier network to contextualize tokens with fair time and space complexity, allowing to process long documents with low-resource hardware and avoid rolling window strategies. Finally, we reach the theoretical computational complexity lower bound for relation extraction using a selective pooling strategy and distance-aware attention weights with trainable polynomial distance functions. The main contributions of this paper are as follows:

\begin{itemize}
  \item We propose a general, lightweight, and efficient model to jointly detect clinical entities and multiple relations, while requiring low computational power and memory, without the use of external tools or hand-crafted rules. The code is available at \href{https://github.com/ds4dh/JNRF}{https://github.com/ds4dh/JNRF}.
  \item We show that this model can be applied to variable length documents, without any architectural changes. More importantly, it has robust performance independent of the document size. 
  \item To the best of our knowledge, this is the first effort to model ADE and medication extraction at the document level. Unlike existing models in the literature, we demonstrate that our approach is able to identify inter-passage relations without the need of window/input size tuning, post-processing or any further engineering.
\end{itemize}

\begin{figure*}[t!]
\centering
\includegraphics[width=1\textwidth]{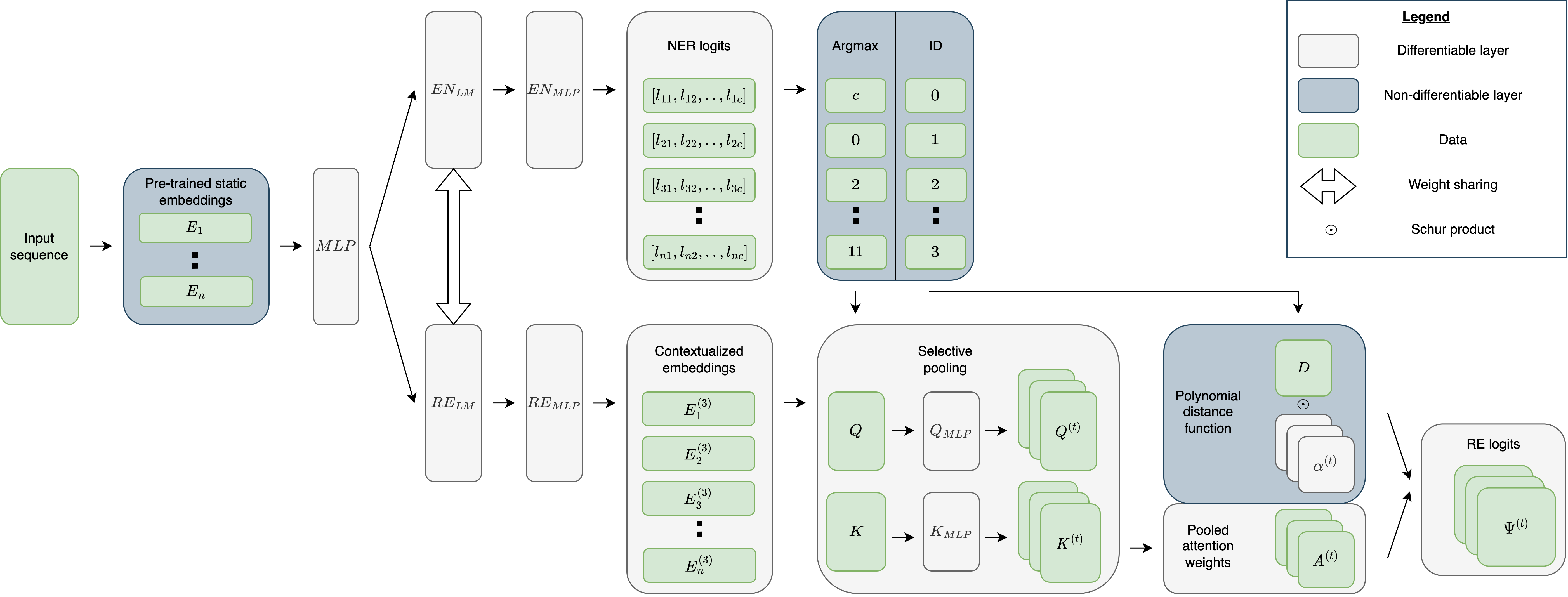}
\caption{Computational graph for the proposed JNRF network.}
\label{fig:JFNet}
\end{figure*}

\section{Related work}
The main methods to produce E2E information extraction systems are the so called \textit{pipeline} \cite{sorokingurevych2017context, chapman2018hybrid, christopoulou2020adverse} and \textit{joint modeling} \cite{xu2017uth_ccb, wei2020study, bekoulis2018joint, nguyen2019end, luanetal2019general, waddenetal2019entity}. The pipeline method consists of training two independent modules, one for NER and one for RE. These models naturally suffer from cascading errors, as the error signal from one module is not back-propagated to the other. Joint modeling aims to overcome this shortcoming by learning a unique model on a combination of NER and RE losses. Joint modeling tends to outperform pipeline methods, consistently achieving state-of-the-art performance \cite{wei2020study,fang2021joint, bekoulis2018joint,nguyen2019end,luanetal2019general,waddenetal2019entity}. In addition, joint modeling techniques have some major advances as they allow to train two models at the same time, saving time and computation, and minimizing engineering efforts. In both cases, the E2E approach has been dominated by LSTM-CRF architectures \cite{xu2017uth_ccb, christopoulou2020adverse, wei2020study, henry20202018}. However, they suffer from two main limitations: \textit{i)} the computational complexity of the CRF layer \cite{jeongetal2009efficient}; and \textit{ii)} the auto-regressive nature of the LSTM model, which prevents full parallel training \cite{xu2021multi}.

\subsection{Joint learning in the general domain}
\citet{bekoulis2018joint} proposed a joint neural model using CRFs and a multi-headed selection module allowing for multiple relation detection. The model requires the computation of scores on every pair of input tokens, which consumes $\mathcal{O}(n^2)$ time and space. To improve generalisation, their approach does not rely on external NLP tools, such as part-of-speech (POS) tagger or dependency parser. More recently, \citet{nguyen2019end} proposed a joint BiLSTM-CRF architecture combined with a biaffine attention mechanism \cite{dozat2016deep}, improving upon \citet{bekoulis2018joint} in terms of time complexity. \citet{luanetal2019general} utilizes dynamic span graphs to learn useful information from a broader context. The graph is built by picking the most confident entity spans and linking them with confidence-weighted relation types and correlations. The model does not require pre-processing syntactic tools and significantly outperforms the previous approaches across several entity-related tasks. Lastly, DYGIE++ \cite{waddenetal2019entity} enumerates candidate text spans and encodes them using BERT and task-specific message updates passed over a text span graph to achieve state-of-the-art performance across entity, relation, and event extraction tasks.

\subsection{Joint learning for medication-related entity and relation extraction}
Most of the medication-related NER and RE studies are performed using the N2C2 ADE benchmark \cite{henry20202018}. \citet{wei2020study} proposed a system consisting of a LSTM-CRF layer for NER joint learned with a CNN-RNN layer for RE. They utilized CLAMP \cite{soysal2018clamp} for the text pre-processing pipeline, including sentence boundary detection and POS labeling, and to extract a set of hand-crafted features to feed the NER module. Similarly to approaches for general corpora, \citet{fang2021joint} replaced the LSTM layer by a BERT model for feature extraction, achieving 1.5 percentage point improvement in the strict F1-score metric. In their approach, a CRF layer is still used on top of a BERT model for the NER part, while a multi-head selection module \cite{bekoulis2018joint} combines the output of the BERT and CRF layers to predict relation among the detected entities.

\subsection{Fourier networks}
To overcome algorithmic complexity limitations in the Transformers architecture \cite{vaswani2017attention}, Fourier networks (FNet) have been proposed \cite{lee2021fnet}. The main innovation of FNets is that the classic Transformers attention mechanism can be mimicked using simple, non-trainable token mixing strategies. One can obtain $\mathcal{O}(n \times \log(n))$ complexity using the Cooley–Tukey Fast Fourier Transform algorithm \citep{ct1965} instead of the attention mechanism, which consumes $\mathcal{O}(n^2)$ with respect to the input sequence length ($n$). FNets achieve 92 and 97\% of BERT-Base and BERT-Large \cite{devlin2018bert} accuracy on the GLUE benchmark \citep{wang2018glue}, but train 70-80\% faster on GPUs/TPUs. In addition to matching the accuracy of competing linear-complexity transformers \citep{wang2020linformer,jaegle2021perceiver,wu2021fastformer,lee2021fnet}, the FNet is faster and memory efficient due to the unparameterized contextualization layer, i.e., it has no parameters to train for token mixing, thus requires virtually no memory usage.

\section{Approach}
In this section, we provide a step-by-step formal description of the proposed architecture using the forward pass representations and operations, as illustrated in Figure \ref{fig:JFNet}. First, we describe \textit{i)} the vectorial token representation strategy, then \textit{ii)} the language/contextualization layer, next \textit{iii)} how the NER and RE task is jointly modelled, and finally \textit{iv)} the cost functions used. Lastly, we conduct a computational complexity analysis of the proposed model.

\subsection{Model formalisation}
\noindent \textbf{Token representation layer:} We use static embeddings (BioClinicalBERT-base \cite{alsentzer2019publicly} in our experiments) and freeze these parameters during training for better generalization. We also decided to use positional encoding as in \citet{vaswani2017attention} so as not to fix a predefined input length.

\par\leavevmode\par

\noindent \textbf{Language model:}  We use FNets to perform token contextualization with fair time and space complexity. We integrate a FNet layer in our architecture as follows:
\begin{align*}
E^{(1)} & = \textrm{MLP}(E), \\ 
E^{(2a)} & = \textrm{EN}_{LM}(E^{(1)}), \\ 
E^{(2b)} & = \textrm{RE}_{LM}(E^{(1)}), 
\end{align*}

\noindent where $E \in \mathbb{R}^{n \times d}$ is the embedding matrix, in which each row represents a token, following their order in the input sequence (i.e., the document), $n$ the input sequence length, $d$ the token embedding dimension, $\textrm{MLP}$ is a token-wise multilayer perceptron, $\textrm{EN}_{LM}$ and $\textrm{RE}_{LM}$ are NER and RE FNets respectively. In fact, we fully share the weights between $\textrm{EN}_{LM}$ and $\textrm{RE}_{LM}$ to further reduce the number of trainable parameters. We use superscripts ($^{(1)}$, $^{(2a)}$, ...) to denote the transformed versions of the original embedding matrix.

\par\leavevmode\par

\noindent \textbf{NER and RE layers:} We thus have $E^{(2)} = E^{(2a)} = E^{(2b)}$, and subsequently compute:
\begin{align*}
l & = \textrm{EN}_{MLP}(E^{(2)}), \\ 
E^{(3)} & = \textrm{RE}_{MLP}(E^{(2)}), 
\end{align*}

\noindent where $\textrm{EN}_{MLP}$ and $\textrm{RE}_{MLP}$ are two independent token-wise MLPs. $\textrm{EN}_{MLP}$ maps the contextualized embeddings $E^{(2)}$ to logits $l \in \mathbb{R}^{n \times c}$ for classification, where $c$ is the number of entity classes, and $\textrm{RE}_{MLP}$ maps $E^{(2)}$ to a third version of the embedding matrix $E^{(3)}$. We then compute a priori token classes
\begin{align*}
a_i & = \textrm{argmax}(l_i), 
\end{align*}

\noindent for $i:1 \; ... \; n$, and apply a selective pooling strategy, i.e., we pool candidate entities for relation extraction from $E^{(3)}$ using $a_i$. Some relations may never exist for a particular relation extraction task. We use $L$ to denote the set of entities that can only be linked to those of a set $H$. To avoid generating impossible candidate pairs, we perform two selective pooling for these two different sets: the key $K \in \mathbb{R}^{|L| \times d}$, and the query $Q \in \mathbb{R}^{|H| \times d}$. We then produce $t$ heads
\begin{align*}
K^{(j)} & = \textrm{K}_{MLP}^{(j)}(K), \\ 
Q^{(j)} & = \textrm{Q}_{MLP}^{(j)}(Q), 
\end{align*}

\noindent for $j:1 \; ... \; t$, where $\textrm{K}_{MLP}^{(j)}$ and $\textrm{Q}_{MLP}^{(j)}$ are token-wise MLPs, and $t$ represent the number of relation types. We then compute the scores between the query and the key entities
\begin{align*}
A^{(j)} & = Q^{(j)} K^{T(j)}.
\end{align*}

\noindent As the RE module is distance agnostic, we incorporate a trainable polynomial distance function to modify the logits as a function of distance between tokens:
\begin{align*}
\Psi^{(j)} & = A^{(j)} + \alpha_{j1} \times D^2 + \alpha_{j2} \times D + \alpha_{j3} \times I, 
\end{align*}

\noindent where $D_{\phi\psi}$ represents the number of tokens separating the $\phi^{th}$ and $\psi^{th}$ pooled entities in the original input embedding matrix. The $\alpha$'s are learned through the minimization of the loss function and thus requires no predefined hand-crafted rules regarding short/long-distance relations.

\par\leavevmode\par

\noindent \textbf{Loss function:} We use a cross-entropy loss for both NER and RE: 
\begin{align*}
\mathcal{L}_{NER} & = - \frac{1}{n}\sum_{i=1}^{n} \sum_{k=1}^{c} s(l_{i, k}) \times e_{i, k}, \\
\mathcal{L}_{RE} & = - \frac{1}{|H||L|}\sum_{h=1}^{|H|}\sum_{p=1}^{|L|} \sum_{j=1}^{t} s(\Psi_{h, p}^{(j)}) \times r_{h, p, j},
\end{align*}

\noindent where $s(x_{q, z}) = \log (\exp \left(x_{q, z}\right)/\sum_{b} \exp \left(x_{q, b}\right))$, and $e$ and $r$ are the target entities and relations, respectively. Finally, we use the sum of $\mathcal{L}_{NER}$ and $\mathcal{L}_{RE}$ as the final loss function to minimize
\begin{align*}
\mathcal{L} & = \mathcal{L}_{NER} + \mathcal{L}_{RE}.
\end{align*}

\subsection{Computational complexity}

\noindent The complexity of the RE model depends on the number of neighbors considered for candidate pair of entities, independently of the method. If one wants to detect relations between two entities regardless of the distance, then the lower bound is $\mathcal{O}(t \times |H| \times |L|)$; or $\operatorname{min}(\mathcal{O}(t \times |L|) \; , \; \mathcal{O}(t \times |H|))$ if one fixes the number of candidate neighbors. We decided not to set a maximum number of neighbors for candidate pair generation. Thus, the RE model uses $\mathcal{O}(t \times |H| \times |L|)$ through selective pooling. For a fixed RE method, the complexity of the whole model is driven by the NER component. We achieved fair complexity by using an FNet ($\mathcal{O}(n \times \log(n))$). Additionally, we used a softmax layer in place of CRF, which uses $\mathcal{O}(n \times c)$ instead of CRF's $\mathcal{O}(n \times c^2)$. This method also takes advantage of parallelization, making it a time complexity optimised method.

\section{Benchmark dataset} \label{Dataset}
We used the 2018 N2C2 ADE dataset \footnote{Dataset available at \href{https://portal.dbmi.hms.harvard.edu/projects/n2c2-2018-t2/}{https://portal.dbmi.hms.harvard.edu/}.} to evaluate our model. The data consists of 505 annotated discharge summaries from MIMIC-III \cite{johnson2016mimic}. The passages contains annotations for \textit{strength}, \textit{form}, \textit{dosage}, \textit{frequency}, \textit{route}, \textit{duration}, \textit{reason}, and \textit{ADE} entities, each associated with a \textit{drug} entity. We used the official splits to train and evaluate our model, with 303 records for training and 202 for testing. Data summary statistics are presented in the Appendix \ref{sec:appendixA}. \textit{Duration} and \textit{ADE} entities and their respective relations are not as well represented in the dataset (see Table \ref{SummaryEntRel}). The document lengths vary widely depending on the patient's clinical history (see Table \ref{SummaryStats}). There is a gap of more than 10k tokens between the smallest and largest documents (224 and 13990, respectively), which is too large to use padding efficiently. Moreover, the average document size is almost 8x larger than the typical input size of standard BERT-like implementations (4045 vs 512, respectively).

\section{Experiments}
We trained our models in three different data representation scenarios, where we use whole documents, sentences only, and a mixed configuration where we use both documents and sentences as training instances. Performance was then evaluated at both document and sentence levels for these different training scenarios. Our models were compared to baseline models based on MLP with selective pooling and a sliding window BioClinicalBERT-base model (WBERT) \citep{alsentzer2019publicly} with selective pooling, both trained and evaluated using the whole documents.

We implemented our models using PyTorch and a single Tesla V100 GPU. We used Adam \cite{kingma2014adam}, mini-batches of size 1 and 64 for documents and sentences, respectively. Models were trained using gradient accumulation to avoid using padding tokens. The final model was selected based on the best dev F1-score obtained during training. In the following, we present the results of our experiments using micro-lenient precision, recall, and F1-score using the challenge's official evaluation tool.

\subsection{Data pre-processing}
We split the provided training data into train and dev sets composed of 242 and 61 documents, respectively. We tokenize documents using BioClinicalBERT-base wordpiece tokenizer from HuggingFace \cite{wu2016google, wolf2019huggingface}. For sentence-level modeling, we first tokenize sentences using Spacy \cite{spacy2} and then use aforementioned wordpiece algorithm. We encode the gold entity boundaries in the BIO scheme. The embedding matrix is initialized from BioClinicalBERT-base static embeddings. No other form a data pre-processing or external feature injection has been implemented.

\subsection{End-to-end effectiveness}
Table \ref{ourscores} shows the performance of the JNRF model in multiple settings. The best performance was obtained in the document-document setting, reaching an end-to-end F1-score of 80.49\%, a precision of 91.65\% and a recall of 71.76\%. The JNRF outperformed WBERT with selective pooling by 0.42\% in F1-score (0.09\% in precision and 0.06\% in recall), while reducing algorithmic complexity by one order of magnitude ($\mathcal{O}(n \times (\log (n) + c))$ \textit{vs.} $\mathcal{O}(n \times (n + c))$). We hypothesize that using WBERT does not improve the performance due to the lack of long-range token mixing and/or an inappropriate windowing strategy. We believe that further investigation of an optimal windowing strategy could improve its performance. Moreover, we observed a significant drop in performance (37\% in F1-score) when the Fnet is replaced by an MLP, demonstrating the capacity of the FNet to better attend to the correct token representations.

The JNRF model shows good performance when it is trained and evaluated with the same document representation (i.e., document-document or sentence-sentence) with similar precision in both cases and reduction in recall for the sentence-sentence setup, due to the model's limitation to detect inter-sentence relations. It is unclear though whether further data engineering could still result in equivalent performance. For the mixed training setup, the model shows stronger power to infer at the sentence level. We believe this is due to the much higher number of examples at the sentence level, which bias the model towards such representation.

\begin{table}[h!]
\centering
\setlength\tabcolsep{1.5pt}
\begin{tabular}{lclrrr}
\hline
\multicolumn{1}{c}{\textbf{Train}} & \multicolumn{1}{c}{\textbf{Language}} & \multicolumn{1}{c}{\textbf{Test}} & \multicolumn{1}{c}{\textbf{Precision}} & \multicolumn{1}{c}{\textbf{Recall}} & \multicolumn{1}{c}{\textbf{F1}} \\

 & \multicolumn{1}{c}{\textbf{model}} &  & \multicolumn{1}{c}{\textbf{(\%)}} & \multicolumn{1}{c}{\textbf{(\%)}} & \multicolumn{1}{c}{\textbf{(\%)}} \\
\hline
 doc. & MLP & doc. & 54.19 & 35.49 & 42.89 \\
\hline
 doc. & WBERT & doc. & 90.66 & 71.70 & 80.07 \\
\hline
\multirow{2}{*}{doc.} & \multirow{2}{*}{FNet} & doc. & \textbf{91.65} & \textbf{71.76} & \textbf{80.49} \\
 & & sent. & 75.28 & 0.29 & 0.57 \\
\hline
 \multirow{2}{*}{sent.} & \multirow{2}{*}{FNet} & doc. & 29.55 & 21.42 & 24.84 \\
 & & sent. & 89.50 & 65.80 & 75.84 \\
\hline
 \multirow{2}{*}{mixed} & \multirow{2}{*}{FNet} & doc. & 66.99 & 32.83 & 44.07 \\
 & & sent. & 81.63 & 62.35 & 70.70 \\
\hline
\end{tabular}
\caption{Lenient micro-averaged E2E scores for different language models and document representations. }
\label{ourscores}
\end{table}

\subsection{End-to-end efficiency}
To compare the efficiency of our approach against architectures used in state-of-the-art approaches, we measured the time and memory used during training over 10 epochs (for the same training set) for a rolling window BERT (WBERT), a rolling window BERT-CRF (WBERT-CRF), and a BiLSTM-CRF. All window-based models used non-overlapping windows of size 512. We deliberately chose to use the minimum number of windows for these models to make them as fast as possible. Figure \ref{fig:TimeSpace} shows the time and VRAM used by our model and state-of-the-art models. Results show that our model substantially improves upon the state-of-the-art in terms of time complexity. Forward and backward passes over the training dataset take an average of 30 seconds with our proposed architecture, while the average time for the above mentioned models is 54, 168 and 685 seconds, respectively. This increases the learning speed by a factor of 2, 6 and 22, respectively (Figure \ref{fig:TimeSpace}a). In addition, we measured an average VRAM usage of 8 GB for the JNRF architecture while the average memory usage for the above mentioned models is 4, 5 and 14 GBs, respectively. This represents a 43\% GPU memory saving compared to BiLSTM-CRF (Figure \ref{fig:TimeSpace}b). WBERT and WBERT-CRF uses around 2x less memory due to the windowing strategy. This increase in efficiency is due to the fact that, differently from the quadratic complexity in terms of the number of entities $c$, which is generally large in the biomedical field, our model complexity has a linear dependency in terms of the number of entities, and a log-linear dependency in terms of the number of tokens (overall $\mathcal{O}(n \times (\log (n) + c))$).

\subsection{Time inefficiency of windowing strategies}
\begin{figure*}[t]\centering
\centering
\includegraphics[width=1\textwidth]{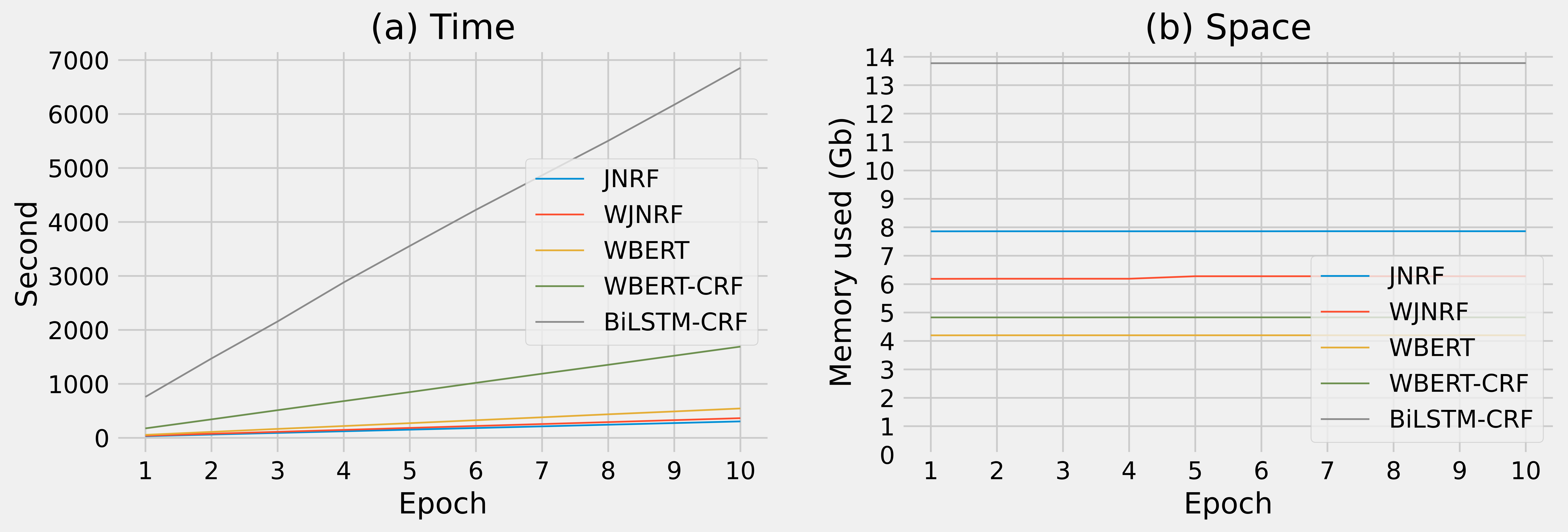}
\caption{(a) Cumulative training time of JNRF \textit{vs.} WJNRF \textit{vs.} WBERT \textit{vs.} WBERT-CRF \textit{vs.} BiLSTM-CRF. (b) GPU memory usage of JNRF \textit{vs.} WJNRF \textit{vs.} WBERT \textit{vs.} WBERT-CRF \textit{vs.} BiLSTM-CRF. For fair comparison, all systems use the selective pooling RE module.}
\label{fig:TimeSpace}
\end{figure*}

To demonstrate that windowing strategies are time inefficient, we measured the average forward-backward time of a rolling window JNRF (WJNRF) and its average VRAM usage (Figure \ref{fig:TimeSpace}). JNRF is 20\% faster than WJNRF but WJNRF uses 26\% less memory (Figure \ref{fig:TimeSpace}). While windowing strategies save VRAM, they are an inefficient solution in terms of computation time. The average document size is 4045 (see Table \ref{SummaryStats}) corresponding to an average of 8 forward passes per document using standard BERT-like implementations (512 tokens maximum input size) or 28 for the longest document. So that all tokens attend to each other, we would need overlapping windows. The worst case scenario is to drag the window token-by-token, leading to 3534 ($n-WindowSize+1$) windows on average per document.

\subsection{Performance across entities, relations and document sizes}
Table \ref{DetailedScores} shows the performance of our model per entity and relation types. Our model suffers from poor performance in extracting \textit{Reason} and \textit{ADE} entities, with an F1-score of 50.26\% and 16.40\%, respectively. This lower performance is also seen in other competing solutions \cite{henry20202018}. In turn, both the detection of their respective relations are also negatively impacted, with a final E2E F1-score of only 29.92\% and 7.21\%, respectively. We believe this lower performance is a result of the confusion between these entities (as they are semantically similar) and of the small number of instances in the training set. Nevertheless, further investigation is needed to better understand the issue. 

\begin{table}[h!]
\centering
\setlength\tabcolsep{3pt}
\begin{tabular}{lccc}
\hline
\multicolumn{1}{c}{\textbf{Entity}} & \multicolumn{1}{c}{\textbf{Precision}} & \multicolumn{1}{c}{\textbf{Recall}} & \multicolumn{1}{c}{\textbf{F1}} \\
\multicolumn{1}{c}{\textbf{}} & \multicolumn{1}{c}{\textbf{(\%)}} & \multicolumn{1}{c}{\textbf{(\%)}} & \multicolumn{1}{c}{\textbf{(\%)}} 
\\ \hline
Drug      & 93.32             & 86.99          & 90.05      \\
Strength  & 96.80             & 95.08          & 95.93      \\
Form      & 96.57             & 92.38          & 94.43      \\
Dosage    & 94.26             & 87.62          & 90.82      \\
Frequency & 97.63             & 92.37          & 94.93      \\
Route     & 92.63             & 93.03          & 92.83      \\
Duration  & 84.98             & 61.38          & 71.27      \\
Reason    & 65.96             & 40.59          & 50.26      \\
ADE       & 36.67             & 10.56          & 16.40      \\
\hline
Overall   & 92.95             & 84.76          & 88.67   \\
\hline
\multicolumn{1}{c}{\textbf{Entity +}} & \multicolumn{1}{c}{\textbf{Precision}} & \multicolumn{1}{c}{\textbf{Recall}} & \multicolumn{1}{c}{\textbf{F1}} \\
\multicolumn{1}{c}{\textbf{Relation}} & \multicolumn{1}{c}{\textbf{(\%)}} & \multicolumn{1}{c}{\textbf{(\%)}} & \multicolumn{1}{c}{\textbf{(\%)}} \\
                        \hline
Strength-Drug  & 95.60             & 88.60          & 91.97      \\
Form-Drug      & 95.63             & 87.01          & 91.12      \\
Dosage-Drug    & 94.13             & 79.07          & 85.94      \\
Frequency-Drug & 94.83             & 83.19          & 88.63      \\
Route-Drug     & 90.50             & 83.25          & 86.72      \\
Duration-Drug  & 76.09             & 41.08          & 53.35      \\
Reason-Drug    & 54.72             & 20.59          & 29.92      \\
ADE-Drug       & 30.30             & 4.09          & 7.21      \\
\hline
Overall                 & 90.97             & 72.08          & 80.43     \\
\hline
\end{tabular}
\caption{\label{DetailedScores}
NER and E2E (NER+RE) performance of our JNRF model.}
\end{table}

Table \ref{ScoreVsLen} shows the performance as a function of the number of input tokens (document length). We followed the Freedman-Diaconis method \cite{freedman1981histogram} to group documents into clusters of different lengths. These results highlights the ability of our architecture to perform consistently across clinical notes of varying sizes. Without any data pre-processing (e.g., sliding window or sentence tokenization), the model can elegantly generalise to document of different sizes.

\begin{table}[h!]
\setlength\tabcolsep{2.75pt}
\begin{tabular}{rrrrr}
\hline
\multicolumn{1}{c}{\textbf{Doc.}} & \multicolumn{1}{c}{\textbf{Doc.}} & \multicolumn{1}{c}{\textbf{Precision}} & \multicolumn{1}{c}{\textbf{Recall}} & \multicolumn{1}{c}{\textbf{F1}} \\ 
\multicolumn{1}{c}{\textbf{length}} & \multicolumn{1}{c}{\textbf{count}} & \multicolumn{1}{c}{\textbf{(\%)}} & \multicolumn{1}{c}{\textbf{(\%)}} & \multicolumn{1}{c}{\textbf{(\%)}} \\ \hline
{[}0, 754{]}                                                                           & 5                                                                                     & 91.67                                 & 59.46                              & 72.13                          \\
{[}754, 1508{]}                                                                        & 8                                                                                     & 97.25                                 & 67.22                              & 79.49                          \\
{[}1508, 2262{]}                                                                       & 18                                                                                    & 89.92                                 & 66.55                              & 76.49                          \\
{[}2262, 3016{]}                                                                       & 28                                                                                    & 91.65                                 & 74.69                              & 82.30                          \\
{[}3016, 3770{]}                                                                       & 43                                                                                    & 91.72                                 & 73.46                              & 81.58                          \\
{[}3770, 4524{]}                                                                       & 30                                                                                    & 90.35                                 & 71.64                              & 79.91                          \\
{[}4524, 5278{]}                                                                       & 32                                                                                    & 90.82                                 & 72.60                              & 80.69                          \\
{[}5278, 6032{]}                                                                       & 18                                                                                    & 89.58                                 & 72.16                              & 79.93                          \\
{[}6032, 6786{]}                                                                       & 10                                                                                    & 93.88                                 & 70.99                              & 80.85                          \\
{[}6786, 7540{]}                                                                       & 4                                                                                     & 89.17                                 & 70.01                              & 78.44                          \\
{[}7540, 8294{]}                                                                       & 3                                                                                     & 88.89                                 & 67.06                              & 76.45                          \\
{[}8294, 9048{]}                                                                       & 1                                                                                     & 92.08                                 & 75.61                              & 83.04                          \\
{[}9802, 10556{]}                                                                      & 1                                                                                     & 88.73                                 & 66.55                              & 76.06                          \\
{[}12064, 12818{]}                                                                     & 1                                                                                     & 92.54                                 & 80.84                              & 86.30                          \\ \hline
\end{tabular}
\caption{Performance of our JNRF model across different document sizes.}
\label{ScoreVsLen}
\end{table}

\subsection{Performance on long range relations}
\begin{figure}[h!]\centering
\centering
\includegraphics[width=1\linewidth]{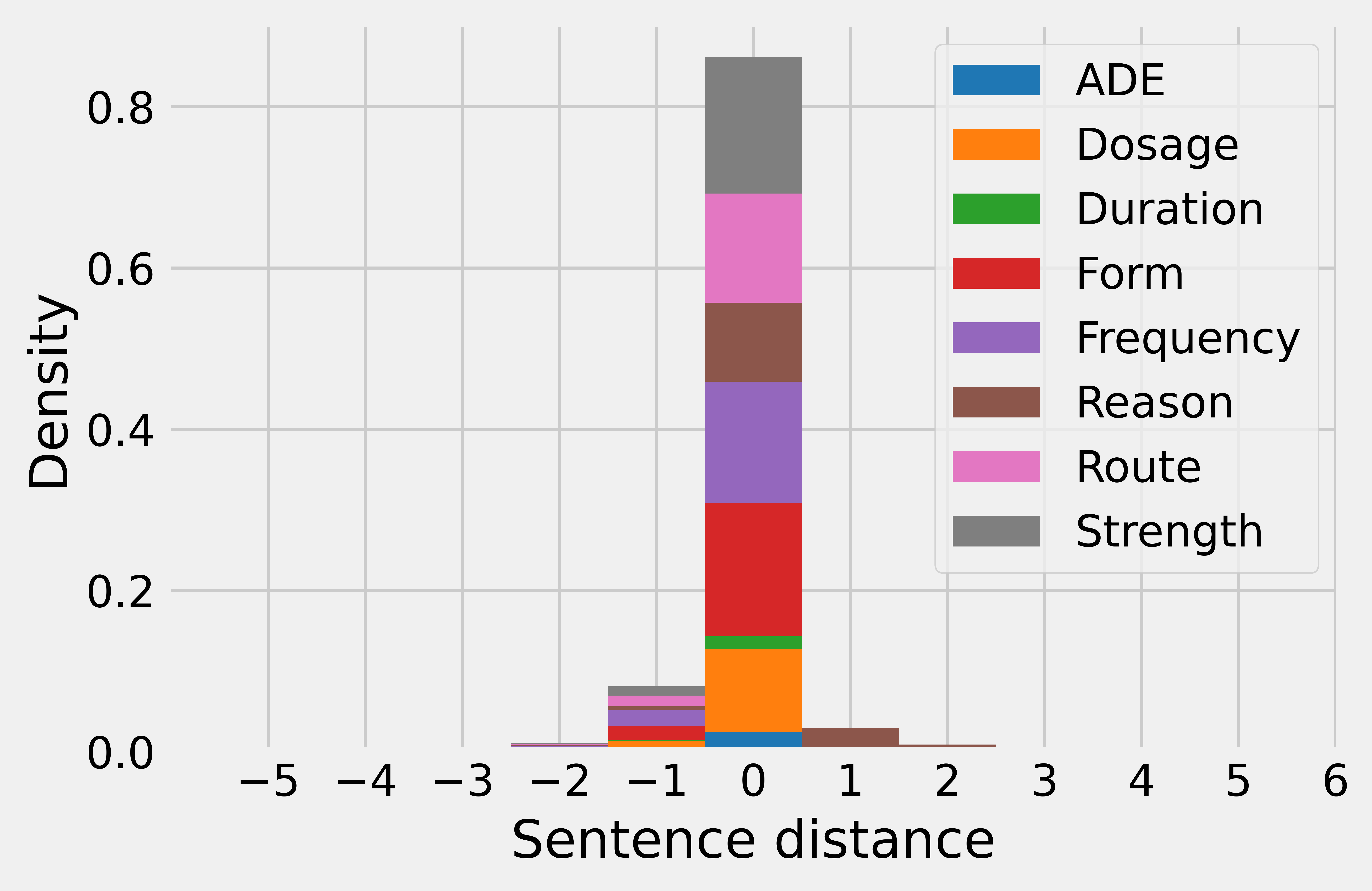}
\caption{Probability density estimation of relation types as a function of the number of sentences separating two related entities (Sentence distance).}
\label{fig:A}
\end{figure}

Figure \ref{fig:A} shows the distribution of relation types according to their sentence distance. We define the sentence distance between two related entities $E1$ and $E2$ as the number of sentences separating $E1$ from $E2$. A negative distance implies that the \textit{drug} entity is mentioned before the related entity. Results show that although most related entities are in the same sentence, there are a non-negligible number of relations with a sentence distance different from zero. As we can see from Table \ref{Doc2DocSentDistMetrics}, the JNRF model is able to automatically detect distant relations. It has superior performance detecting intra-sentence relations, i.e., better F1-score for sentence distance 0, with a yet robust performance for inter-sentence relations with negative sentence distances (between 65\% and 68\% F1-score). The performance decreases substantially for inter-sentence relations with positive sentence distances. This is due to the fact that \textit{Reason} and \textit{ADE} entities and relations are actually harder to detect (see Table \ref{DetailedScores}), and they represent the vast majority of relations with a positive sentence distance, as shown in Figure \ref{fig:A}. It is important to note that using a fixed-input size models would only detect intra-sentence relations or inter-sentence through significant engineering, which may not necessarily generalise to other corpora and domains.

\begin{table}[h!]
\centering
\setlength\tabcolsep{3pt}
\begin{tabular}{lrrrrr}
\hline
    & \multicolumn{5}{c}{\textbf{Sentence distance}}                \\
    & \multicolumn{1}{c}{\textbf{-2}} & \multicolumn{1}{c}{\textbf{-1}} & \multicolumn{1}{c}{\textbf{0}} & \multicolumn{1}{c}{\textbf{1}} & \multicolumn{1}{c}{\textbf{2}} \\
\hline
\textbf{Precision (\%)} & 75.14 & 83.06 & 92.69 & 22.99 & 0.36 \\
\textbf{Recall (\%)} & 56.90 & 57.88 & 76.08 & 5.82 & 0.49 \\
\textbf{F1-score (\%)} & 64.76 & 68.22 & 83.57 & 9.29 & 0.41 \\
\hline
\end{tabular}
\caption{Performance of our JNRF model as a function of sentence distance.}
\label{Doc2DocSentDistMetrics} 
\end{table}

\section{Comparison with SOTA in the N2C2 ADE challenge}
In this section, for a reference we show our results against state-of-the-art E2E NER+RE models described in the N2C2 ADE challenge \cite{henry20202018}. Nevertheless, due to their different modelling strategy (e.g., multiple models, external tools, post-processing techniques and hand-crafted rules specifically designed for this dataset), they are not directly comparable.

\par\leavevmode\par

\textbf{UTH} \citep{wei2020study} used a joint learning model consisting of a LSTM-CRF layer for NER and a CNN-RNN layer for RE. CLAMP \cite{soysal2018clamp} was employed for text pre-processing, including sentence boundary detection and POS labeling, and to create a set of hand-crafted features that fed the CRF layer. Entities without a relation were associated to the closest drug in the post-processing step.

\textbf{NaCT} \citep{christopoulou2020adverse} used a majority voting ensemble of feature-based CRF, including ADE dictionary, and stacked BiLSTM-CRF for NER. For RE, they used an ensemble of LSTM for intra-sentence relations and a transformer network for inter-sentence relations.

\textbf{BCH} \citep{miller2019extracting} used SVM to detect entities, and pair these detected entities for a second SVM relation classifier. They used cTAKES \cite{savova2010mayo} to pre-process data and ClearTK \cite{bethardetal2014cleartk} API to extract features.

\textbf{RA} \citep{henry20202018} used dictionary-based features, CRFs and logistic regression for NER. For RE, they used a tree-based boosting classifier \cite{chen2016xgboost}.

\par\leavevmode\par

Table \ref{n2c2scores} shows the performance of our best model as well as the results of the previously described systems. As we can see, the performance of our E2E model (80.49\% F1-score) achieves 90\% of the F1-score of the best performing system (99\% precision and 84\% recall), while significantly reducing algorithmic complexity. Moreover, it compares favorably to strong baseline methods \cite{chen2016xgboost} (80.49\% \textit{vs.} 80.37\%), again with an order of magnitude in complexity reduction.

\begin{table}[h!]
\centering
\setlength\tabcolsep{2.5pt}
\begin{tabular}{lllllrrr}
\hline
 \multicolumn{1}{c}{\textbf{Name}}  &  \multicolumn{1}{c}{\textbf{NER}} &  \multicolumn{1}{c}{\textbf{Precision}} &  \multicolumn{1}{c}{\textbf{Recall}} &  \multicolumn{1}{c}{\textbf{F1}}\\
  &  \multicolumn{1}{c}{\textbf{complexity}} &  \multicolumn{1}{c}{\textbf{(\%)}} &  \multicolumn{1}{c}{\textbf{(\%)}} &  \multicolumn{1}{c}{\textbf{(\%)}}\\
\hline
UTH   & $ n  c^2$ & \textbf{92.92} & \textbf{85.49} & \textbf{89.05} \\
NaCT  & $ n  c^2$ &  92.64 & 83.18 & 87.66 \\
BCH & $n^3$ &  89.63 & 76.40 & 82.49 \\
JNRF  & $n  (\log(n)+c)$ &  91.65$^a$ & 71.76$^b$ & 80.49$^c$  \\
RA & $n  c^2$ &  86.89 & 74.75 & 80.37  \\
\hline
\end{tabular}
\caption{E2E scores of the top performing systems submitted in the N2C2 ADE track, along with our JNRF model. Standard deviations: a=0.47, b=0.53, c=0.33.}
\label{n2c2scores}
\end{table}

\section{Conclusion}
In this paper, we proposed an end-to-end, generalizable, lightweight, and efficient model to jointly detect entities and multiple relations at the intra- and inter-passage levels. We combined a Fourier network with a pooled attention layer to significantly reduce time and space complexity, thus providing the community with a low carbon footprint solution for end-to-end relation extraction. We demonstrated that our model outperformed the sliding window BERT with selective pooling by 0.42\% in F1-score, while being 2 times faster to train. Furthermore, we showed that our model trains 22 times faster and consumes 1.75 times less GPU memory than state-of-the-art BiLSTM-CRF architectures, with a reasonable performance tradeoff of 90\% on the N2C2 ADE benchmark, without using external tools or hand-crafted rules. Furthermore, we showed that this approach achieves consistent performance regardless of the length of the input sequence, eliminating the need for sliding window techniques and easing the overall data processing pipeline and engineering effort.

\bibliography{anthology,custom}
\bibliographystyle{acl_natbib}

\newpage

\appendix
\section{Appendix}
\subsection{N2C2 dataset summary statistics} \label{sec:appendixA}

\begin{table}[h!]
\setlength\tabcolsep{3pt}
\begin{tabular}{lrrr}
\hline
\multicolumn{1}{c}{\textbf{Entity type}}          & \multicolumn{1}{c}{\textbf{Full (\%)}} & \multicolumn{1}{c}{\textbf{Training}} & \multicolumn{1}{c}{\textbf{Test}} \\ \hline
Drug                     & 26.8k (32)                       & 16.2k                                & 10.6k                            \\
Strength                                   & 10.9k (13)                       & 6.7k                                  & 4.2k                              \\
Form                                       & 11.0k (13)                       & 6.7k                                  & 4.4k                              \\
Dosage                                     & 6.9k (8)                          & 4.2k                                  & 2.7k                              \\
Frequency                                  & 10.3k (12)                       & 6.3k                                  & 4.0k                              \\
Route                                      & 9.0k (11)                         & 5.5k                                  & 3.5k                              \\
Duration                                   & 1.0k (1)                           & 0.6k                                   & 0.4k                               \\
Reason                                     & 6.4k (8)                          & 3.9k                                  & 2.5k                              \\
ADE                                        & 16k (2)                          & 1.0k                                   & 0.6k                               \\ \hline
Total                                      & 83.8k (100)                      & 51.0k                                & 32.9k                            \\ \hline
\multicolumn{1}{c}{\textbf{Relation type}} & \multicolumn{1}{c}{\textbf{Full (\%)}} & \multicolumn{1}{c}{\textbf{Training}} & \multicolumn{1}{c}{\textbf{Test}} \\ \hline
Strength-Drug                              & 10.9k (18)                       & 6.7k                                  & 4.2k                              \\
Form-Drug                                  & 11.0k (19)                       & 6.7k                                  & 4.4k                              \\
Dosage-Drug                                & 6.9k (11)                         & 4.2k                                  & 2.7k                              \\
Frequency-Drug                             & 10.3k (17)                       & 6.3k                                 & 4.0k                              \\
Route-Drug                                 & 9.1k (15)                         & 5.5k                                  & 3.5k                              \\
Duration-Drug                              & 1.1k (2)                          & 0.6k                                   & 0.4k                               \\
Reason-Drug                                & 8.6 (15)                         & 5.2k                                  & 3.4k                              \\
ADE-Drug                                   & 1.8 (3)                          & 1.1k                                  & 0.7k                               \\ \hline
Total                                      & 59.8 (100)                      & 36.4k                                & 23.5k                            \\ \hline
\end{tabular}
\caption{Entity and relation distributions.}
\label{SummaryEntRel}
\end{table}

\begin{table}[h!]\centering
\begin{tabular}{lrrr}
                      \hline
              & \textbf{Train set } & \textbf{Validation set} & \textbf{Test set} \\
                      \hline
\textbf{Count} & 242            & 61           & 202           \\
\textbf{Mean}  & 4045           & 3829         & 3933          \\
\textbf{Std}   & 1972           & 1870         & 1790          \\
\textbf{Min}   & 224            & 237          & 252           \\
\textbf{Max}   & 13990          & 7845         & 12518        \\
                      \hline
\end{tabular}
\caption{Statistics of document length in terms of tokens.}
\label{SummaryStats}
\end{table}

\end{document}